\documentclass[10pt,twocolumn,letterpaper]{article}

\usepackage[table,dvipsnames]{xcolor}

\usepackage[pagenumbers]{iccv} 

\definecolor{iccvblue}{rgb}{0.21,0.49,0.74}
\usepackage[pagebackref,breaklinks,colorlinks,allcolors=iccvblue]{hyperref}

\usepackage{multirow}
\usepackage{pifont}
\usepackage{algorithm}
\usepackage[noend]{algpseudocode}
\algdef{SE}[DOWHILE]{Do}{DoWhile}{\algorithmicdo}[1]{\algorithmicwhile\ #1}
\algnewcommand\algorithmicforeach{\textbf{for each}}
\algdef{S}[FOR]{ForEach}[1]{\algorithmicforeach\ #1\ \algorithmicdo}

\newcommand{\cc}{\cellcolor{gray!40}}

\def\paperID{7937} 
\def\confName{ICCV}
\def\confYear{2025}

\title{NegRefine: Refining Negative Label-Based Zero-Shot OOD Detection}

\author{Amirhossein Ansari$^{1}$ \quad Ke Wang$^{1}$ \quad Pulei Xiong$^{2}$ \vspace{0.3em} \\
{\normalsize $^1$Simon Fraser University} \quad\quad
{\normalsize $^2$National Research Council Canada} \\
{\tt\small \nolinkurl{ah_ansari@sfu.ca}, \nolinkurl{wangk@cs.sfu.ca}, \nolinkurl{pulei.xiong@nrc-cnrc.gc.ca}}
}

\begin{document}
\maketitle

\begin{abstract}
Recent advancements in Vision-Language Models like CLIP have enabled zero-shot OOD detection by leveraging both image and textual label information. Among these, negative label-based methods such as NegLabel and CSP have shown promising results by utilizing a lexicon of words to define negative labels for distinguishing OOD samples. However, these methods suffer from detecting in-distribution samples as OOD due to negative labels that are subcategories of in-distribution labels or proper nouns. They also face limitations in handling images that match multiple in-distribution and negative labels. We propose NegRefine, a novel negative label refinement framework for zero-shot OOD detection. By introducing a filtering mechanism to exclude subcategory labels and proper nouns from the negative label set and incorporating a multi-matching-aware scoring function that dynamically adjusts the contributions of multiple labels matching an image, NegRefine ensures a more robust separation between in-distribution and OOD samples. We evaluate NegRefine on large-scale benchmarks, including ImageNet-1K. The code is available at {\small \url{https://github.com/ah-ansari/NegRefine}}.
\end{abstract}

\section{Introduction}
\label{sec:intro}
Machine learning models deployed in real-world environments often encounter input samples from unknown classes that differ significantly from the pre-defined classes in the training data. Model predictions for these Out-of-Distribution (OOD) samples are typically unreliable and overconfident, posing critical risks and security vulnerabilities in critical domains like healthcare and autonomous driving \cite{nguyen2015deep}. OOD detection mitigates this issue by distinguishing OOD samples from in-distribution ones, enabling models to reject and flag OOD inputs when encountered \cite{hendrycks17baseline,yang2024generalized,ansari2024out}. 
Traditionally, image OOD detection methods have relied on single-visual modality \cite{liang2018enhancing,liu2020energy,wei2022logitnorm,du2022vos}, neglecting the textual information available in class labels. However, with recent advancements in pre-trained Vision-Language Models like CLIP \cite{radford2021learning}, adopting a multi-modal paradigm that also incorporates class label information has gained significant attention \cite{esmaeilpour2022zero, ming2022delving, wang2023clipn}. Among these, negative label-based methods, such as NegLabel \cite{jiang2024neglabel} and CSP \cite{chen2024conjugated}, have emerged as particularly promising solutions.


The core idea behind NegLabel \cite{jiang2024neglabel} is to leverage a large lexicon of English words, such as WordNet \cite{fellbaum1998wordnet}, as a label pool to create a set of negative (OOD) labels $Y_{neg}$. To form $Y_{neg}$, words (nouns and adjectives) from the lexicon are ranked based on their semantic similarity to the in-distribution labels in $Y_{in}$, and the top $p\%$ least similar words to the in-distribution labels are selected as negative labels. For an input image, the similarity between the image and both the in-distribution labels in $Y_{in}$ and the negative labels in $Y_{neg}$ is computed using CLIP, and based on these similarity scores, a final score is defined to detect the image as in-distribution or OOD. CSP \cite{chen2024conjugated} improves upon NegLabel by refining the use of adjectives to construct superclass labels that better capture OOD samples.

Despite the impressive performance of NegLabel \cite{jiang2024neglabel} and CSP \cite{chen2024conjugated}, these methods still face the following notable limitations:

\begin{figure*}[h!]
  \centering
  \begin{subfigure}{0.212\linewidth}
    \includegraphics[width=\linewidth]{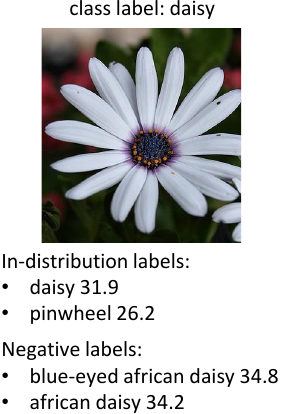}
    \caption{Subcategory Overlap}
    \label{fig:subcategory}
  \end{subfigure}
  \hfill
  \begin{subfigure}{0.212\linewidth}
    \includegraphics[width=\linewidth]{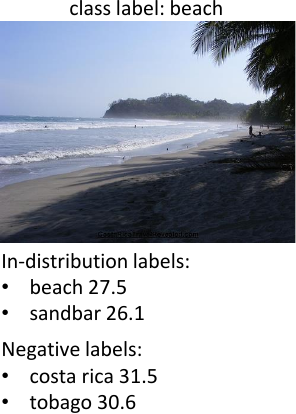}
    \caption{Proper Noun}
    \label{fig:proper-noun}
  \end{subfigure}
  \hfill
  \begin{subfigure}{0.212\linewidth}
    \includegraphics[width=\linewidth]{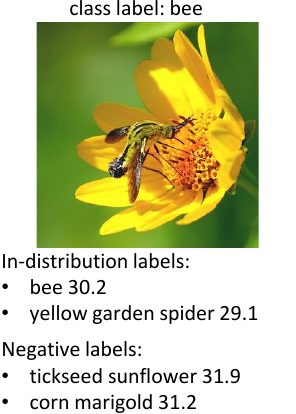}
    \caption{Multi-Label Matching}
    \label{fig:mm-multi-object}
  \end{subfigure}
  \hfill
  \begin{subfigure}{0.212\linewidth}
    \includegraphics[width=\linewidth]{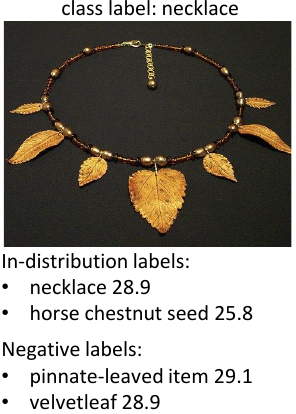}
    \caption{Multi-Label Matching}
    \label{fig:mm-single-object}
  \end{subfigure}  
  \caption{In-distribution images from ImageNet-1K with the top two labels from $Y_{in}$ and $Y_{neg}$. The value beside each label represents CLIP's similarity score multiplied by 100. These images are incorrectly detected as ``OOD'' by CSP due to: (a) a subcategory of the in-distribution label appearing in the negative labels, (b) proper nouns in the negative labels receiving high similarity scores. (c) and (d) multi-label matching issue, with negative matching labels receiving higher similarity scores.}
  \label{fig:issues}
\end{figure*}

\textbf{Subcategory Overlap.} 
These methods rely on textual similarity and a cutoff of the top $p\%$ least similar words to select $Y_{neg}$. This approach is inherently limited, as it depends solely on semantic similarity without explicitly considering hierarchical relationships among words. As a result, they can mistakenly include subcategories (\eg, ``blue-eyed african daisy'') of in-distribution labels (\eg, ``daisy'') in the negative set, leading to an overlap between $Y_{neg}$ and $Y_{in}$. This issue is particularly pronounced because, for a given image, CLIP tends to assign higher similarity scores to more specific labels than to general terms, consequently, the negative subcategory label can receive a higher similarity score than the more general in-distribution label, leading to incorrect detection of in-distribution samples as OOD. \cref{fig:subcategory} shows this issue using an in-distribution sample from the ImageNet-1K \cite{deng2009imagenet} dataset.

\textbf{Proper Nouns.} 
WordNet contains many proper nouns (\eg, names of individual persons, places, or organizations), leading to their excessive selection as negative labels. The CLIP model tends to match samples from certain in-distribution classes with proper nouns in $Y_{neg}$ with very high similarity, resulting in their incorrect detection as OOD. \cref{fig:proper-noun} shows an image from the in-distribution class ``beach'', for which CLIP assigns a much higher similarity score to some proper nouns in $Y_{neg}$, such as ``costa rica'' (a country) and ``tobago'' (an island), than to the true label ``beach'' from $Y_{in}$, even though the image is not actually from ``costa rica'' or ``tobago''. One likely reason is that these locations are well known for their beaches or strongly associated with beach images in CLIP's training data.

\textbf{Multi-Label Matching.} 
A real-world image often contains multiple objects \cite{miyai2025gl}, matching the labels of all the objects in the image. Even a single-object image can match multiple labels, with each label describing a different aspect of the image. In such cases, if an image from an in-distribution class also matches some negative labels, it may be incorrectly detected as OOD by NegLabel and CSP. This is because these methods rely solely on the significance of CLIP’s similarity scores for individual labels in $Y_{in}$ and $Y_{neg}$, without accounting for multiple labels matching an image. \cref{fig:mm-multi-object,fig:mm-single-object} illustrate this issue using two in-distribution images: (c) a multi-object image containing both a ``bee'' (in-distribution label) and a flower ``tickseed sunflower'' (negative label), and (d) a single-object image of a ``necklace'' (in-distribution label) with the shape of a leaf, matching ``pinnate-leaved item'' (negative label). These images are incorrectly detected as OOD by CSP because the negative labels received higher similarity scores.

In this paper, we propose NegRefine, a novel framework that addresses the limitations of existing negative label-based methods for zero-shot OOD detection. Our contributions are as follows:

\begin{itemize}
    \item NegRefine incorporates a filtering mechanism to prevent the inclusion of subcategory labels and proper nouns in the negative label set, ensuring a clearer semantic separation between $Y_{in}$ and $Y_{neg}$.

    \item NegRefine introduces a novel scoring function that accounts for multiple labels matching an image, dynamically adjusting their contributions based on contextual relevance to mitigate the misprediction of in-distribution images as OOD when they also match negative labels.
    
    \item We extensively evaluate NegRefine on large-scale benchmarks, including ImageNet-1K. As noted in \cite{bitterwolf2023ninco}, a significant portion of existing OOD datasets belongs to one of the classes in ImageNet-1K, making them unsuitable as true OOD samples. Following \cite{bitterwolf2023ninco}, we adopt more reliable OOD datasets for ImageNet-1K. Our method outperforms the state-of-the-art CSP \cite{chen2024conjugated} by 1.82\% increase in AUROC and 4.35\% reduction in FPR95.

\end{itemize}

\section{Preliminaries}

\subsection{Problem Definition}
\label{sec:problem}
Given in-distribution class labels $Y_{in} = \{y_1, y_2, \dots, y_K\}$, where $y_i$ is the textual label for class $i$ and $K$ is the number of classes, CLIP-based zero-shot OOD detection \cite{esmaeilpour2022zero, ming2022delving, wang2023clipn,jiang2024neglabel} is to detect whether an input image $x$ is ``in-distribution'' (\ie, belongs to one of the classes in $Y_{in}$) or ``OOD'', using the CLIP model and $Y_{in}$ as input. In the zero-shot setting, no training data is available to fine-tune CLIP or adjust its text prompt. Solving this problem requires designing an in-distribution scoring function $S(x)$, which assigns higher scores to in-distribution samples and lower scores to OOD samples. Given a threshold $\gamma$, an image $x$ is detected as in-distribution if $S(x) \geq \gamma$, and as OOD otherwise.


\subsection{CLIP Model}
\label{sec:clip}
CLIP \cite{radford2021learning} is a vision-language model trained on a large corpus of (image, caption) pairs using a contrastive loss to align images with their corresponding text representations in a shared embedding space. It consists of an image encoder, $f_{\mathrm{img}}:x \rightarrow \mathbb{R}^D$, and a text encoder, $f_{\mathrm{txt}}:t \rightarrow \mathbb{R}^D$, which map an image $x$ and a text $t$ to a $D$-dimensional feature space. In standard CLIP-based zero-shot classification, a text prompt is first created for each class label $y_i \in Y$ in a format like ``This is a $y_i$'' (\eg, ``This is a dog''). The cosine similarity is then computed between the input image embedding, $f_{\mathrm{img}}(x)$, and each text prompt embedding, $f_{\mathrm{txt}}(\text{``This is a } y_i\text{''})$. The image $x$ is assigned to the class label with the highest cosine similarity.


\subsection{Negative Label-Based OOD Detection}
\label{sec:neglabel}
NegLabel \cite{jiang2024neglabel} introduced creating a set of negative (OOD) labels, $Y_{neg}$, to fully leverage CLIP for zero-shot OOD detection. To select the negative labels, nouns and adjectives from WordNet were ranked based on their similarity to the in-distribution labels, computed using CLIP’s text encoder. The top $p=15\%$ least similar words to the in-distribution labels were then selected as the negative label set $Y_{neg}$. Using $Y_{in}$ and $Y_{neg}$, they defined their in-distribution score as:
\begin{equation} 
\label{eq:neg-label} 
    S_{NegLabel}(x) = \frac{ \sum_{i=1}^{K} e^{\operatorname{sim}(x, y_i)/\tau} }{ \sum_{i=1}^{K} e^{\operatorname{sim}(x, y_i)/\tau} + \sum_{j=1}^{M} e^{\operatorname{sim}(x, \tilde{y}_j)/\tau} },
\end{equation}
where $y_i$ represents the $i$-th in-distribution label (from $K$ in-distribution labels), and $\tilde{y}_j$ is the $j$-th negative label (from $M$ negative labels). The function $\operatorname{sim}(x,y_i)$ denotes the cosine similarity between the embedding of the image $x$ and the embedding of the text for label $y_i$. The temperature parameter is set to $\tau=0.01$ to scale the similarity scores. CSP \cite{chen2024conjugated} follows the same structure as NegLabel but enhances it by leveraging adjectives from WordNet to generate modified superclass labels, which better match OOD samples.

\section{Methodology}
\label{sec:method}
We propose NegRefine to address the limitations of current negative label-based methods discussed in the Introduction, \ie, the inclusion of subcategories of in-distribution labels and proper nouns in the negative set, and the inability to effectively handle images matching multiple labels. NegRefine has two key components: (i) a negative label filtering mechanism to improve the quality of negative labels, and (ii) an effective multi-matching score that dynamically adjusts to handle in-distribution images matching multiple labels. The outcome is a new score function $S$ based on the refined negative label set produced by (i) and the adjustment applied by (ii). With this new score, the OOD detection is the same as described in \cref{sec:problem}. In the rest of this section, we discuss the negative label filtering mechanism and the multi-label matching score function.

\subsection{Negative Label Filtering Mechanism}
\label{sec:negfilter}
As discussed, subcategory labels like ``african daisy'' receive higher similarity than their more general in-distribution labels (\cref{fig:subcategory}), and proper nouns like ``costa rica'' can match certain in-distribution images (\cref{fig:proper-noun}), leading to incorrect detection of such images as OOD. To mitigate these issues, we propose to remove such negative labels. Note that removing proper nouns does not impact OOD detection, as proper nouns are so specific that if an OOD image is truly related to a proper noun, a more general non-proper noun negative label (\eg, ``island'' instead of ``tobago'') will likely capture the image. This is because a good negative set is expected to cover all OOD images.

Our negative label filtering method, NegFilter, is presented in \cref{alg:negfilter}. It starts with the initial set of negative labels selected in CSP \cite{chen2024conjugated}, denoted as $Y_{neg}$, and refines it by filtering out improper negative labels, \ie, proper nouns and subcategories. To identify such labels, we leverage an off-the-shelf Large Language Model (LLM) to automate the filtering process. LLMs have demonstrated near-human performance in various NLP tasks \cite{naveed2023comprehensive,chang2024survey}, making them a reliable tool for this purpose. For each term $w$ in $Y_{neg}$, we apply the following two filtering checks:
\begin{itemize}
    \item \textbf{Proper Nouns Check:} We query the LLM with the prompt ``Is $w$ a proper noun, like the name of an entity?'' (line 3). If the response is ``Yes'', we remove $w$ from $Y'_{neg}$ (line 5). 
    
    \item \textbf{Subcategory Check:} We first identify the set $L$ of the top $n$ most similar labels in $Y_{in}$ to $w$ using CLIP’s text encoder (lines 7–12) to minimize the number of subcategory queries. Then, for each $l \in L$, we query the LLM with the prompt: ``Is $w$ a subcategory of $l$?''. If the LLM confirms that $w$ is a subcategory of any label in this set, we remove $w$ from $Y'_{neg}$ (line 11).
\end{itemize}
The resulting negative label set $Y_{neg}'$ has lower similarity to in‐distribution images, reducing the likelihood of misdetecting in-distribution samples as OOD.

WordNet \cite{fellbaum1998wordnet} also provides a hierarchy of words that can be used to identify subcategories. However, we observed several issues that make WordNet unreliable for our task: ``african daisy'' and ``daisy'' are at the same level instead of forming a parent-child pair; closely related words are placed in different branches, such as some subcategories of ``mushroom'' fall under ``fungus'', while others fall under ``fungus-genus'' (fungus is the general term for mushrooms).  Moreover, WordNet cannot be used to identify proper nouns. Therefore, we opted to use LLMs for their higher accuracy, greater flexibility, and ease of use.

\begin{algorithm}[tb]
  \caption{NegFilter($Y_{neg}, Y_{in}, LLM, f_{\mathrm{txt}}, n$)}
  \label{alg:negfilter}
  \textbf{Input}: $Y_{neg}$ (initial negative label set), $Y_{in}$ (in-distribution label set), $LLM$ (Large Language Model), $f_{\mathrm{txt}}$ (CLIP text encoder), $n$ (number of in-distribution labels for subcategory check).\\  
  \textbf{Output}: $Y'_{neg}$ (refined negative label set). 
  \begin{algorithmic}[1]
    \State $Y'_{neg} \gets Y_{neg}$
    \ForEach {$w \in Y_{neg}$}
        \State $R \gets LLM($ ``Is $w$ a proper noun, like the name of an entity?''$)$
        \If{$R =$ ``Yes''}
            \State $Y'_{neg} \gets Y'_{neg} - \{w\}$
        \Else
            \State $L \gets$ $n$ top similar labels to $w$ in $Y_{in}$ using $f_{\mathrm{txt}}$
            \ForEach {$l \in L$}
                \State $R \gets LLM($ ``Is $w$ a subcategory of $l$?''$)$
                \If{$R =$ ``Yes''}
                    \State $Y'_{neg} \gets Y'_{neg} - \{w\}$
                    \State \textbf{break}
                \EndIf
            \EndFor
        \EndIf
    \EndFor
    \State \textbf{return} $Y'_{neg}$
  \end{algorithmic}
\end{algorithm}

\subsection{Multi-Label Matching Score Function}
\label{sec:smm}
The in-distribution score used in NegLabel and CSP (\cref{eq:neg-label}) relies on the similarity scores of individual in-distribution and negative labels, but it overlooks cases where in-distribution images also match negative labels with high similarity (see \cref{fig:mm-multi-object,fig:mm-single-object}). To address this issue, we propose a new score that can adjust and capture such multi-label matching cases.

Consider the training paradigm of the CLIP model. CLIP is trained on (image, caption) pairs, where captions often describe multiple objects or aspects of an image. This training enables CLIP to develop a deep understanding of an image’s visual content. Motivated by this, instead of relying on individual label similarity as in CSP and NegLabel, we propose creating a more caption-like text by concatenating an in-distribution label $y$ with a negative label $\tilde{y}$, forming the text ``$y$ and $\tilde{y}$''. For example, for the image in \cref{fig:mm-multi-object}, the concatenated text is ``bee and tickseed sunflower''. If the similarity score for this text significantly increases compared to the score for the negative label ``tickseed sunflower'' alone, it suggests that the joint description better captures the image content and reinforces the relevance of the in-distribution label ``bee''. Conversely, if the similarity remains unchanged or decreases, it suggests that the labels do not jointly describe the image, and so the in-distribution label does not contribute.

To illustrate, consider the images in \cref{fig:mm-multi-object,fig:mm-single-object}, which truly match both in-distribution and negative labels:
\begin{itemize}
    \item \cref{fig:mm-multi-object}: the similarity for ``bee'' (in-distribution) is 30.2, and for ``tickseed sunflower'' (negative) is 31.9, while the similarity increases for ``bee and tickseed sunflower'' to 34.4.
    \item \cref{fig:mm-single-object}: the similarity for ``necklace'' (in-distribution) is 28.9, and for ``pinnate-leaved item'' (negative) is 29.1, while the similarity increases for ``necklace and pinnate-leaved item'' to 34.0.
\end{itemize}
These examples show how similarity significantly increases for the concatenated text when the two labels truly correspond to the image.

To formalize our approach, let $T_{in}$ and $T_{neg}$ denote the top $k$ in-distribution and negative labels that best match an image based on CLIP similarity scores. For each $y_i \in T_{in}$ and $\tilde{y}_j \in T_{neg}$, we create the text $t_{i,j}$ as ``$y_i$ and $\tilde{y}_j$''. With $k$ in-distribution and $k$ negative labels, this approach generates $k^2$ text combinations. We define a multi-matching score $S_{MM}$ to measure the contribution of the in-distribution label $y_i$ through the comparison between the concatenated text and the negative label alone. For an input image $x$, $S_{MM}(x)$ is defined as:
\begin{equation}  
\label{eq:smm}  
    S_{MM}(x) = \max_{i, j} \; \frac{ e^{\operatorname{sim}( x, t_{i,j})/\tau} }{ e^{\operatorname{sim}( x, t_{i,j})/\tau} + e^{\operatorname{sim}(x, \tilde{y}_j)/\tau} },  
\end{equation}  
where $\operatorname{sim}(x, t_{i,j})$ denotes the cosine similarity between the embeddings of the image $x$ and text $t_{i,j}$.  

For a multi-matching in-distribution image, the in-distribution label $y_i$ is truly relevant to the image, thus, the similarity of $t_{i,j}$ (``$y_i$ and $\tilde{y}_j$'') will be higher than that of $\tilde{y}_j$ alone, resulting in a larger $S_{MM}$ score. Conversely, for an OOD image, the in-distribution label $y_i$ is not relevant, so the similarity of $t_{i,j}$ will be equal to or lower than that of $\tilde{y}_j$, leading to a smaller $S_{MM}$ score.

\noindent \textbf{Final Score.} 
The initial in-distribution score from NegLabel, $S_{NegLabel}$ (see \cref{eq:neg-label}), applies a softmax and summation over individual in-distribution and negative label similarity scores to measure the overall attraction toward in-distribution or OOD. In contrast, we propose $S_{MM}$ as an additional adjustment to capture multi-matching cases. We combine the collective summation from $S_{NegLabel}$ with the multi-matching adjustment from $S_{MM}$ to define our final in-distribution score as:
\begin{equation}
\label{eq:final-score}
    S(x) = S_{NegLabel}(x) + \alpha \times S_{MM}(x).
\end{equation}
Here, $\alpha$ is a weighting factor that controls the contribution of the multi-matching score $S_{MM}$ to the final score.

\section{Experiments}

\begin{table*}
  \centering
  \resizebox{\textwidth}{!}{
  \begin{tabular}{@{}lcccccccc|cc@{}}
    \toprule
    & \multicolumn{8}{c}{OOD Datasets} & \\
    \multirow{2}{*}{Methods} & \multicolumn{2}{c}{iNaturalist} & \multicolumn{2}{c}{OpenImage-O} & \multicolumn{2}{c}{Clean} & \multicolumn{2}{c}{NINCO} & \multicolumn{2}{|c}{Average} \\ \cmidrule(lr){2-3} \cmidrule(lr){4-5} \cmidrule(lr){6-7} \cmidrule(lr){8-9} \cmidrule(lr){10-11}
    & AUROC $\uparrow$ & FPR95 $\downarrow$ & AUROC $\uparrow$ & FPR95 $\downarrow$ & AUROC $\uparrow$ & FPR95 $\downarrow$ & AUROC $\uparrow$ & FPR95 $\downarrow$ & AUROC $\uparrow$ & FPR95 $\downarrow$ \\
    \midrule
    ZOC \cite{esmaeilpour2022zero} & 79.53 & 90.98 & 80.61 & 70.04 & 72.88 & 81.07 & 67.96 & 78.67 & 75.24 & 80.19 \\
    MCM \cite{ming2022delving} & 94.59 & 32.20 & 92.00 & 41.03 & 83.24 & 57.79 & 74.34 & 79.33 & 86.04 & 52.59 \\
    GL-MCM \cite{miyai2025gl} & 96.44 & 17.42 & 92.91 & 34.52 & 84.78 & 51.49 & 76.03 & 74.40 & 87.54 & 44.46 \\
    CLIPN \cite{wang2023clipn} & 96.20 & 19.11 & 92.22 & 30.36 & 87.31 & 41.56 & 78.72 & 66.51 & 88.61 & 39.39  \\
    NegLabel \cite{jiang2024neglabel} & 99.49 & 1.91 & 93.74 & 28.62 & 86.79 & 41.22 & 77.30 & 68.70 & 89.33 & 35.11 \\
    CSP \cite{chen2024conjugated} & \textbf{99.60} & 1.54 & 94.09 & 28.94 & 88.32 & 38.75 & 77.88 & 68.65 & 89.97 & 34.47 \\ \midrule
    NegRefine (ours) & 99.57 & \textbf{1.47} & \textbf{95.00} & \textbf{23.85} & \textbf{90.65} & \textbf{33.18} & \textbf{81.92} & \textbf{61.96} & \textbf{91.79} & \textbf{30.12} \\
    \bottomrule
  \end{tabular}}
  \caption{OOD detection performance on the ImageNet-1K in-distribution benchmark. The average is over the four OOD datasets. All methods are zero-shot and utilize CLIP ViT-B/16 model. The metrics are reported as percentages, with the best results highlighted in bold.}
  \label{tab:main-result}
\end{table*}

\noindent \textbf{Datasets and Benchmarks.} 
The main evaluation in most previous studies \cite{jiang2024neglabel, chen2024conjugated, wang2023clipn} has been conducted using the ImageNet-1K \cite{deng2009imagenet} dataset as the in-distribution data, with subsets of iNaturalist \cite{van2018inaturalist}, SUN \cite{xiao2010sun}, Places \cite{zhou2017places}, and the full Textures \cite{cimpoi2014describing} dataset, used as OOD \cite{huang2021mos}. However, \cite{bitterwolf2023ninco} randomly selected 400 samples from 12 commonly used OOD datasets and, through manual examination, found that 59.5\% of the samples from Places and 25.6\% from Textures were actually in-distribution (\ie, they could be assigned to one of the classes in ImageNet-1K). Among all 12 datasets analyzed, only iNaturalist \cite{van2018inaturalist} (2.5\%) and OpenImage-O \cite{wang2022vim} (4.9\%) exhibited significantly lower in-distribution rates. We extended this analysis to the SUN dataset, which was not covered in \cite{bitterwolf2023ninco}, and observed an in-distribution rate of 26.2\%.
To address this high in-distribution rate, the same study \cite{bitterwolf2023ninco} introduced the NINCO dataset as a more reliable OOD data for ImageNet-1K and also provided the set of clean OOD samples obtained from their manual analysis of 400 samples across the 12 datasets, which we refer to as Clean. In our evaluation, we consider NINCO and Clean and also include iNaturalist and OpenImage-O, which contain minimal in-distribution contamination. Thus, our OOD datasets for the ImageNet-1K benchmark are iNaturalist, OpenImage-O, Clean, and NINCO. A description of these OOD datasets is provided in \cref{sec:supp-ood-data}.

\noindent \textbf{Implementation Details.} 
We employ the CLIP ViT-B/16 \cite{radford2021learning} as the pre-trained CLIP model. For the initial negative labels, label prompts for CLIP, and other parameter details, we followed the settings of CSP \cite{chen2024conjugated}. For our negative label filtering mechanism (\cref{alg:negfilter}), we use Qwen2.5-14B-Instruct \cite{bai2023qwen} as the LLM and set $n=10$. For the multi-matching score $S_{MM}$, we use $k=5$ and set $\alpha=2$ as the weight of $S_{MM}$ in \cref{eq:final-score}. All experiments are conducted on NVIDIA RTX 6000 GPUs.

\noindent \textbf{Evaluation Metrics.} 
Following previous studies \cite{du2022vos, esmaeilpour2022zero, wang2023clipn}, we evaluate our method using the Area Under the Receiver Operating Characteristic Curve (AUROC) and the False Positive Rate at the 95\% True Positive Rate, denoted by FPR95. In these measures, in-distribution data is treated as the positive class.

\noindent \textbf{Baselines.} 
We compare our method, NegRefine, with state-of-the-art CLIP-based zero-shot OOD detection baselines listed in the first column of \cref{tab:main-result}. See Related Work for details on these baselines. In this comparison, we exclude few-shot and supervised training methods, as they require training data, whereas we focus on the zero-shot setting.

\begin{table*}
  \centering
  \resizebox{\textwidth}{!}{
  \begin{tabular}{@{}cc|cccccccc|cc@{}}
    \toprule
    \multirow{2}{*}{$S_{MM}$} & \multirow{2}{*}{NegFilter} & \multicolumn{2}{c}{iNaturalist} & \multicolumn{2}{c}{OpenImage-O} & \multicolumn{2}{c}{Clean} & \multicolumn{2}{c}{NINCO} & \multicolumn{2}{|c}{Average} \\ \cmidrule(lr){3-4} \cmidrule(lr){5-6} \cmidrule(lr){7-8} \cmidrule(lr){9-10} \cmidrule(lr){11-12}
    & & AUROC $\uparrow$ & FPR95 $\downarrow$ & AUROC $\uparrow$ & FPR95 $\downarrow$ & AUROC $\uparrow$ & FPR95 $\downarrow$ & AUROC $\uparrow$ & FPR95 $\downarrow$ & AUROC $\uparrow$ & FPR95 $\downarrow$ \\ 
    \midrule
    \ding{53} & \ding{53} & 99.60 & 1.54 & 94.09 & 28.94 & 88.32 & 38.75 & 77.88 & 68.65 & 89.97 & 34.47 \\ \midrule
    \ding{53} & \checkmark & \textbf{99.66} & \textbf{1.28} & 94.92 & 24.93 & 89.41 & 35.37 & 79.83 & 65.93 & 90.95 & 31.88 \\
    $\checkmark $ & \ding{53} & 99.46 & 1.74 & 94.52 & 26.23 & 89.99 & 35.26 & 81.42 & 63.57 & 91.35 & 31.70 \\ \midrule
    \checkmark & \checkmark & 99.57 & 1.47 & \textbf{95.00} & \textbf{23.85} & \textbf{90.65} & \textbf{33.18} & \textbf{81.92} & \textbf{61.96} & \textbf{91.79} & \textbf{30.12} \\ 
    \bottomrule
  \end{tabular}}
  \caption{Analysis of the two main components of our method. \ding{53} denotes that the component is turned off, and $\checkmark$ indicates it is applied. The first row, \ie all components are off, corresponds to CSP.}
  \label{tab:main-ablation}
\end{table*}

\begin{table*}
  \centering
  \resizebox{\textwidth}{!}{
  \begin{tabular}{rc|cccccccc|cc@{}}
    \toprule
    \multicolumn{2}{c|}{\;\;NegFilter\;\;} & \multicolumn{2}{c}{iNaturalist} & \multicolumn{2}{c}{OpenImage-O} & \multicolumn{2}{c}{Clean} & \multicolumn{2}{c}{NINCO} & \multicolumn{2}{|c}{Average} \\ \cmidrule(lr){1-2} \cmidrule(lr){3-4} \cmidrule(lr){5-6} \cmidrule(lr){7-8} \cmidrule(lr){9-10} \cmidrule(lr){11-12} 
    \;\;C1 & C2 & AUROC $\uparrow$ & FPR95 $\downarrow$ & AUROC $\uparrow$ & FPR95 $\downarrow$ & AUROC $\uparrow$ & FPR95 $\downarrow$ & AUROC $\uparrow$ & FPR95 $\downarrow$ & AUROC $\uparrow$ & FPR95 $\downarrow$ \\ 
    \midrule
    \ding{53} & \ding{53} & 99.46 & 1.74 & 94.52 & 26.23 & 89.99 & 35.26 & 81.42 & 63.57 & 91.35 & 31.70 \\ \midrule
    \checkmark & \ding{53} & 99.51 & 1.62 & 94.78 & 25.15 & 90.40 & 34.34 & 81.63 & 63.27 & 91.58 & 31.09 \\
    \ding{53} & \checkmark & 99.54 & 1.63 & 94.78 & 24.81 & 90.29 & 34.06 & 81.65 & 62.80 & 91.56 & 30.82 \\ \midrule
    \checkmark & \checkmark & \textbf{99.57} & \textbf{1.47} & \textbf{95.00} & \textbf{23.85} & \textbf{90.65} & \textbf{33.18} & \textbf{81.92} & \textbf{61.96} & \textbf{91.79} & \textbf{30.12} \\ 
    \bottomrule
  \end{tabular}}
  \caption{Analysis of the filtering checks used in our negative label filtering component. C1 refers to the proper nouns check, and C2 refers to the subcategory check. $S_{MM}$ is always applied.}
  \label{tab:prompt-ablation}
\end{table*}

\subsection{Main Results}
\cref{tab:main-result} presents the results comparing the performance of our method (NegRefine) against previous CLIP-based zero-shot OOD detection studies on the ImageNet-1K benchmark.  Our method outperforms the state-of-the-art approach, CSP, achieving a 1.82\% increase in average AUROC and a 4.35\% reduction in average FPR95. More specifically, out of the eight AUROC and FPR95 metrics evaluated across the four OOD datasets, our method achieves the best result in seven cases, while CSP only slightly outperforms us by 0.03\% on AUROC for iNaturalist. 
Results for our method are averaged over 10 different seeds, with standard deviations reported in \cref{tab:main-std}. For details about the implementation of baselines and the source of randomness in our method and other baselines, refer to \cref{sec:supp-baseline-implementation,sec:supp-randomness}.

\subsection{Analysis}

\noindent \textbf{Analysis of Main Components.} 
Our method introduces two key components: (i) NegFilter (\cref{alg:negfilter}) to remove negative labels that are proper nouns and subcategories of in-distribution labels, and (ii) the $S_{MM}$ score (\cref{eq:smm}) to  handle in-distribution images matching multiple labels. \cref{tab:main-ablation} presents an ablation study analyzing the individual contribution of each component. The first row corresponds to the CSP method, serving as the reference for comparison. Applying negative label filtering alone improves the average FPR95 by 2.59\%, while incorporating $S_{MM}$ alone reduces FPR95 by 2.77\%. Integrating both components in our final method (last row) results in a 4.35\% reduction in average FPR95.

\noindent \textbf{Analysis of Individual Filters in NegFilter.} 
In \cref{tab:prompt-ablation}, we analyze the contributions of the two filtering checks used in our negative label filtering (NegFilter): proper nouns filtering, denoted by C1, and subcategory filtering, denoted by C2. $S_{MM}$ is always applied as the base case in this study. C1 alone reduces the average FPR95 by 0.61\%, while C2 reduces it by 0.88\%. When applied together, they achieve a combined reduction of 1.58\% in average FPR95.

The proper nouns check removes 1749 negative labels, while the subcategory check removes 307 negative labels, accounting for 20.6\% and 3.6\% of the initial set of negative labels, respectively. Examples of removed proper nouns are costa rica, tobago, golden gate bridge, budapest, and saint christopher, and examples of removed subcategories are african daisy (daisy), morchella (mushroom), calendula (daisy), tandoor (stove), with the corresponding supercategory in-distribution labels shown in parentheses. The presence of these erroneous negative labels can lower the in-distribution score for in-distribution samples, leading to misdetection, as discussed in \cref{fig:subcategory,fig:proper-noun}.

\begin{figure*}
  \centering
  \begin{subfigure}{0.325\linewidth}
    \includegraphics[width=\linewidth]{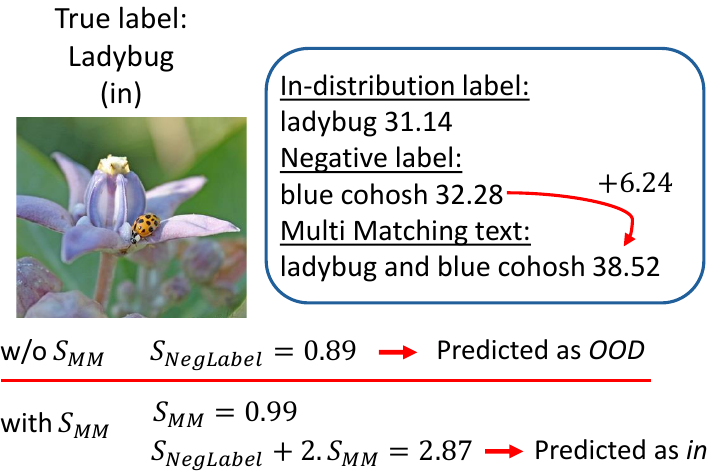}
    \caption{}
    \label{fig:smm-a}
  \end{subfigure}
  \hfill
  \begin{subfigure}{0.325\linewidth}
    \includegraphics[width=\linewidth]{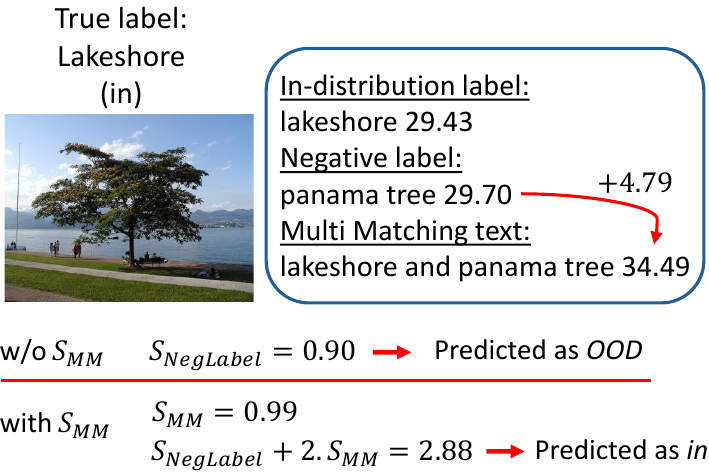}
    \caption{}
    \label{fig:smm-b}
  \end{subfigure}
  \hfill
  \begin{subfigure}{0.325\linewidth}
    \includegraphics[width=\linewidth]{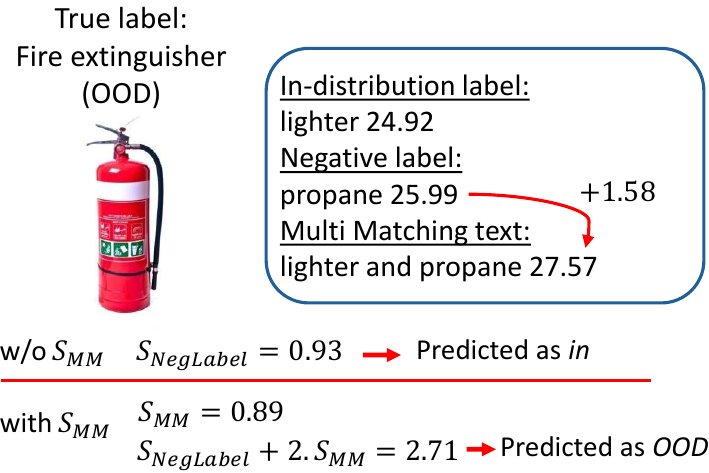}
    \caption{}
    \label{fig:smm-c}
  \end{subfigure}
  \caption{Effect of incorporating $S_{MM}$. The detection threshold $\gamma$ (set such that 95\% of in-distribution data are detected as in-distribution) is 0.91 for $S_{NegLabel}$ and 2.85 for $S$. (a) and (b) are in-distribution images from ImageNet-1K, and (c) is an OOD sample from NINCO. The text boxes display the top labels and multi-matching texts with their CLIP similarity scores ($\times100$). Using $S_{NegLabel}$, all three predictions are incorrect due to multi-matching issue. By incorporating $S_{MM}$, the final score $S$ is adjusted, leading to correct predictions.}
  \label{fig:smm}
\end{figure*}

\noindent \textbf{Analysis of the Effect of $S_{MM}$.} 
\cref{fig:smm} illustrates how incorporating $S_{MM}$ improves OOD detection using two in-distribution examples  (a) and (b) from ImageNet-1K and an OOD example (c) from the NINCO dataset. Using the original score $S_{NegLabel}$, all three examples are predicted incorrectly due to the multi-label matching issue. By applying $S_{MM}$, that is, using the new score $S = S_{NegLabel} + 2 S_{MM}$, all samples are correctly predicted. The reason for this improvement lies in how $S_{MM}$ behaves for different types of samples. For the in-distribution images (a) and (b), the change in similarity from the negative label alone to the multi-matching text is significant, 6.24 in (a) and 4.79 in (b), resulting in a high $S_{MM}$ score. This large change is because, for these examples, both in-distribution and negative labels match the image, making the combined text a better description for the content of the image (refer to \cref{sec:smm}). Conversely, for the OOD sample (c), this change is smaller, only 1.58, leading to a low $S_{MM}$ score. In this way, $S_{MM}$ adjusts the final score to correct the prediction.

\noindent \textbf{Comparing $S_{MM}$ to CLIP’s Local Features.} 
Recently, GL-MCM \cite{miyai2025gl} proposed using CLIP’s local embeddings alongside its global (final) embedding for OOD detection to address images with multiple objects, a motivation similar to our $S_{MM}$ score. In CLIP ViT-B/16, these local embeddings are the $14\times14$ patch tokens, while the global embedding refers to the final [CLS] token output \cite{zhou2022extract}. Their intuition is that local embeddings better capture local objects, thus improving detection for images with multiple objects. \cref{tab:main-result} showed that our method outperforms GL-MCM.

\begin{figure}[t]
  \centering
  \includegraphics[width=0.8\linewidth]{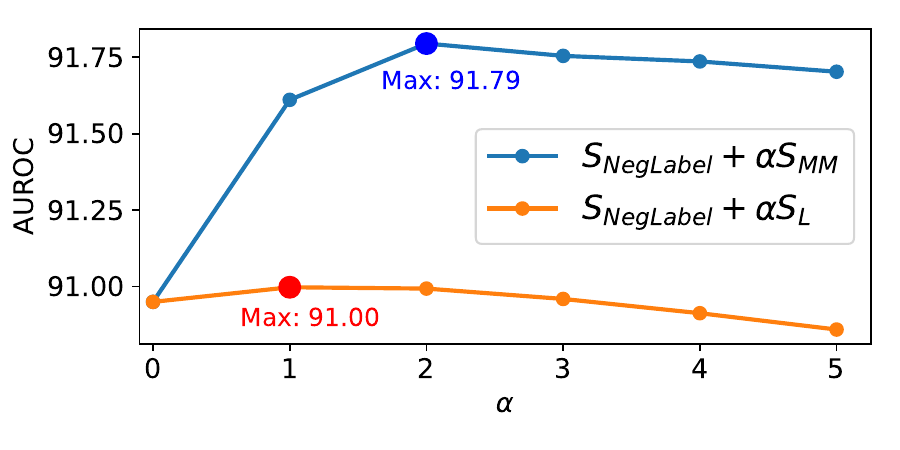}
   \caption{Comparison of our $S_{MM}$ with the local-features-based approach $S_{L}$ for addressing the multi-matching issue. AUROC $\uparrow$, averaged over the OOD datasets across different $\alpha$, is reported.}
   \label{fig:local-features}
\end{figure}

For a more direct comparison with our $S_{MM}$, we extend their local features idea to the negative label-based score by replacing the global image embedding with local embeddings in \cref{eq:neg-label}. We compute the score for each of the $14\times14$ local embeddings and take the maximum as the local score $S_{L}(x)$. Using this, we define a new in-distribution score $S'(x) = S_{NegLabel}(x) + \alpha S_{L}(x)$. We compare $S'(x)$ with our score $S(x)=S_{NegLabel}(x) + \alpha S_{MM}(x)$ in \cref{fig:local-features}. For a fair comparison, negative label filtering is applied for both scores. 
Our $S$ outperforms $S'$ across all $\alpha$ values. This superior performance can be attributed to two main reasons: First, $S_{MM}$ addresses more general multi-label matching cases; for example, even single-object images can match multiple labels (see \cref{fig:mm-single-object}), where the local-features approach would be ineffective. Second, our score, which uses the global embedding, aligns more closely with CLIP’s training, as the global embedding is explicitly optimized to match the corresponding caption.

\noindent \textbf{Extended Analysis and Ablations.} 
Further analyses and ablations on the size of the initial negative label set, the choice of LLM in NegFilter, different CLIP architectures, and alternative designs and parameters for $S_{MM}$ are provided in \cref{sec:supp-ablations}.

\begin{table*}
  \centering
  \resizebox{\textwidth}{!}{
  \begin{tabular}{@{}lcccccccccc@{}}
    \toprule
    in-distribution data & \multicolumn{2}{c}{ImageNet-10} & \multicolumn{2}{c}{ImageNet-10} & \multicolumn{2}{c}{ImageNet-20} & \multicolumn{2}{c}{ImageNet-99} & \multicolumn{2}{c}{WaterBirds} \\
    OOD data & \multicolumn{2}{c}{ImageNet-20} & \multicolumn{2}{c}{ImageNet-99} & \multicolumn{2}{c}{ImageNet-10} & \multicolumn{2}{c}{ImageNet-10} & \multicolumn{2}{c}{Placesbg} \\ \midrule
    Methods & AUROC $\uparrow$ & FPR95 $\downarrow$ & AUROC $\uparrow$ & FPR95 $\downarrow$ & AUROC $\uparrow$ & FPR95 $\downarrow$ & AUROC $\uparrow$ & FPR95 $\downarrow$ & AUROC $\uparrow$ & FPR95 $\downarrow$ \\
    \midrule
    NegLabel & 98.80 & 5.00 & 99.45 & 1.89 & 98.04 & 11.60 & 89.53 & 52.89 & 87.99 & 29.16 \\
    CSP & \textbf{99.02} & \textbf{3.30} & \textbf{99.50} & \textbf{1.78} & 98.79 & 3.40 & 91.21 & 44.35 & 92.88 & 12.07 \\ \midrule
    NegRefine (ours) & 98.95 & 3.89 & 99.21 & 2.24 & \textbf{99.19} & \textbf{2.28} & \textbf{91.68} & \textbf{40.99} & \textbf{93.13} & \textbf{11.96} \\
    \bottomrule
  \end{tabular}}
  \caption{OOD detection performance on ImageNet subsets and the Spurious OOD (WaterBirds vs. Placesbg).}
  \label{tab:imagenet-subset}
\end{table*}

\begin{table*}
  \centering
  \resizebox{\textwidth}{!}{
  \begin{tabular}{@{}llcccccccc|cc@{}}
    \toprule
    \multirow{2}{*}{in-distribution data} & \multirow{2}{*}{Methods} & \multicolumn{2}{c}{iNaturalist} & \multicolumn{2}{c}{OpenImage-O} & \multicolumn{2}{c}{Clean} & \multicolumn{2}{c}{NINCO} & \multicolumn{2}{|c}{Average} \\ \cmidrule(lr){3-4} \cmidrule(lr){5-6} \cmidrule(lr){7-8} \cmidrule(lr){9-10} \cmidrule(lr){11-12}
    && AUROC $\uparrow$ & FPR95 $\downarrow$ & AUROC $\uparrow$ & FPR95 $\downarrow$ & AUROC $\uparrow$ & FPR95 $\downarrow$ & AUROC $\uparrow$ & FPR95 $\downarrow$ & AUROC $\uparrow$ & FPR95 $\downarrow$ \\
    \midrule
    \multirow{3}{*}{ImageNet-Sketch}
    & NegLabel & 99.34 & 2.24 & 93.04 & 30.46 & 85.42 & 43.35 & 74.78 & 70.30 & 88.15 & 36.59 \\
    & CSP & \textbf{99.49} & \textbf{1.60} & \textbf{93.74} & 30.86 & 87.68 & 40.96 & 76.91 & 69.87 & 89.45 & 35.82 \\
    & NegRefine & 99.25 & 1.95 & 93.60 & \textbf{29.27} & \textbf{89.05} & \textbf{37.42} & \textbf{79.90} & \textbf{66.52} & \textbf{90.45} & \textbf{33.79} \\ \midrule
    \multirow{3}{*}{ImageNetV2}
    & NegLabel & 99.40 & 2.47 & 92.54 & 31.86 & 84.92 & 44.57 & 74.01 & 71.37 & 87.72 & 37.57 \\
    & CSP & \textbf{99.54} & \textbf{1.76} & 93.06 & 31.96 & 86.80 & 41.55 & 74.57 & 71.13 & 88.49 & 36.60 \\
    & NegRefine & 99.51 & 1.77 & \textbf{94.25} & \textbf{26.38} & \textbf{89.57} & \textbf{34.95} & \textbf{79.96} & \textbf{64.36} & \textbf{90.82} & \textbf{31.87} \\ \midrule
    \multirow{3}{*}{ImageNet-A}
    & NegLabel & 98.80 & 4.09 & 86.56 & 42.33 & 82.78 & 50.61 & 66.31 & 79.33 & 83.61 & 44.09 \\
    & CSP & \textbf{99.15} & \textbf{2.91} & 88.33 & 43.62 & 86.36 & 46.48 & 70.69 & 78.21 & 86.13 & 42.80 \\
    & NegRefine & 98.86 & 3.45 & \textbf{89.17} & \textbf{40.40} & \textbf{87.75} & \textbf{42.98} & \textbf{74.48} & \textbf{74.02} & \textbf{87.56} & \textbf{40.21} \\ \midrule
    \multirow{3}{*}{ImageNet-R}
    & NegLabel & 99.58 & 1.60 & \textbf{97.02} & \textbf{15.22} & 94.75 & \textbf{21.36} & \textbf{88.74} & \textbf{50.88} & \textbf{95.02} & \textbf{22.27} \\
    & CSP & \textbf{99.79} & \textbf{0.89} & 96.47 & 18.15 & 94.17 & 22.80 & 85.83 & 54.59 & 94.06 & 24.11 \\
    & NegRefine & 99.68 & 0.98 & 96.75 & 16.59 & \textbf{94.77} & 21.47 & 87.08 & 51.85 & 94.57 & 22.72 \\
    \bottomrule
  \end{tabular}}
  \caption{OOD detection performance on domain-shifted versions of ImageNet.}
  \label{tab:domain-shift}
\end{table*}

\subsection{Additional Benchmarks}
\noindent \textbf{ImageNet Subsets and Spurious OOD.}
Following \cite{jiang2024neglabel, chen2024conjugated}, we also use subsets of ImageNet-1K, namely ImageNet-10, ImageNet-20, and ImageNet-100, created in \cite{ming2022delving} for evaluation. These subsets contain 10, 20, and 100 classes, respectively. However, we modify ImageNet-100 by removing the ``race car'' class, resulting in ImageNet-99. This modification is because we observed that ``race car'' includes samples identical to those in ``sports car'' from ImageNet-10, creating conflicts when evaluating ImageNet-10 (in) against ImageNet-100 (OOD). Additionally, we evaluate our method on the Spurious OOD benchmark \cite{ming2022impact}, where WaterBirds \cite{sagawa2019distributionally} (in) is tested against Placesbg (OOD). Results for these benchmarks are presented in \cref{tab:imagenet-subset}.

In this study, when using ImageNet-10 as in-distribution data, our performance is slightly lower than CSP and NegLabel. This is because, with only 10 classes, this subset has high separability between in-distribution and OOD data, making the challenges we address less significant and limiting the potential for further improvement. However, our method outperforms CSP and NegLabel in all other cases, particularly in the more challenging scenario of ImageNet-99 (in) vs. ImageNet-10 (OOD), where FPR95 is largely higher for all methods.

\noindent \textbf{Robustness to Domain Shift.} 
In this benchmark, we use domain-shifted versions of ImageNet as the in-distribution data, including ImageNet-Sketch \cite{wang2019learning}, ImageNetV2 \cite{recht2019imagenet}, ImageNet-A \cite{hendrycks2021natural}, and ImageNet-R \cite{hendrycks2021many}. The first two datasets share the same class labels as ImageNet-1K, while the class labels in the other two are subsets of ImageNet-1K. Since our method relies solely on in-distribution class labels, we use the same set of negative labels as for ImageNet-1K in the first two datasets. For the other datasets, we reselect the negative label set based on their in-distribution labels. The results are presented in \cref{tab:domain-shift}. In this experiment, our method outperformed CSP and NegLabel on most datasets, except for ImageNet-R, where NegLabel performed slightly better than our method.

\noindent \textbf{Other In-Distribution Datasets.} 
Evaluations using other datasets as in-distribution data, including Stanford-Cars \cite{krause20133d}, CUB-200 \cite{wah2011caltech}, Oxford-Pet \cite{parkhi2012cats}, and Food-101 \cite{bossard2014food}, are provided in \cref{sec:supp-other-in-distribution}.

\section{Related Work}
In this section, we review previous studies that utilized CLIP for OOD detection.

\noindent \textbf{Zero-Shot Methods.} 
ZOC \cite{esmaeilpour2022zero} develops a caption generator to suggest unseen labels. It then uses CLIP's similarity for in-distribution and unseen labels to define a detection score. MCM \cite{ming2022delving} applies softmax over CLIP similarity for in-distribution labels and uses the maximum softmax probability as the in-distribution score. CLIPN \cite{wang2023clipn} introduces a ``No'' text encoder to assist CLIP in understanding prompts like ``not containing $y_i$'', and uses this for OOD detection. GL-MCM \cite{miyai2025gl} improves MCM by leveraging CLIP’s local features to address images containing multiple objects. This is a motivation similar to that of our $S_{MM}$. However, the multi-matching issue we address is more general than the presence of multiple objects in an image, as even single-object images can match multiple labels (see \cref{fig:mm-single-object}).

NegLabel \cite{jiang2024neglabel} introduces negative label-based OOD detection by mining negative labels from WordNet \cite{fellbaum1998wordnet} and defining an in-distribution score based on CLIP’s similarity to in-distribution and negative labels. CSP \cite{chen2024conjugated} improves upon NegLabel by introducing a conjugated semantic pool, where adjectives from WordNet are combined with a predefined set of general labels (\eg, area, creature, item) to generate modified superclass labels. These labels can more effectively capture OOD samples. We build on these two studies by addressing their limitations, primarily to reduce the misdetection of in-distribution samples as OOD.

LAPT \cite{zhang2024lapt} and AdaNeg \cite{zhang2024adaneg} build on NegLabel by applying additional techniques: LAPT uses prompt tuning to optimize NegLabel's text prompts, while AdaNeg employs a test-time adaptation strategy. Our work is orthogonal to these two studies, as we improve the internal mechanism of NegLabel. Notably, LAPT and AdaNeg can also be applied on top of our work by replacing NegLabel with our NegRefine to further enhance performance.

\noindent \textbf{Few-Shot and Full-Data Methods.} 
These methods rely on training data (to a different extent) and are therefore not directly related to ours, as we focus on the zero-shot setting. NPOS \cite{tao2023nonparametric} fine-tunes CLIP’s image encoder and generates OOD samples in the feature space to refine the in-distribution/OOD boundary. LoCoOp \cite{miyai2023locoop} optimizes prompts by leveraging irrelevant parts of the in-distribution data as OOD signals to enhance in-distribution/OOD separation. NegPrompt \cite{li2024learning} learns negative prompts with semantics opposite to in-distribution classes and then performs OOD detection based on similarity to the learned negative prompts. LSN \cite{nie2024out} learns class-specific negative prompts that capture features absent in the in-distribution classes. SeTAR \cite{li2024setar} enhances OOD detection by applying low-rank approximations to the weight matrices of CLIP.



\section{Conclusion}
We introduced NegRefine, a novel framework for zero-shot OOD detection, addressing key limitations of previous negative-label-based methods. NegRefine integrates a negative label filtering mechanism and a multi-matching-aware scoring function. The filtering mechanism refines the negative label set by removing subcategory labels and proper nouns, ensuring that the negative set has lower similarity to in-distribution samples. The scoring function dynamically adjusts the contribution of in-distribution labels, mitigating the incorrect detection of in-distribution samples as OOD caused by the multi-label matching issue. Extensive evaluations on ImageNet-1K benchmark with reliable OOD data demonstrate that NegRefine outperforms state-of-the-art methods on most OOD datasets.

\section*{Acknowledgments}
This project was supported in part by a discovery grant for Ke Wang from Natural Sciences and Engineering Research Council of Canada, and by collaborative research funding from the National Research Council of Canada’s Artificial Intelligence for Logistics Program.


{
    \small
    \bibliographystyle{ieeenat_fullname}
    \bibliography{main}
}

\clearpage

\appendix

\section{Extended Experiments}

\subsection{OOD Dataset Descriptions}
\label{sec:supp-ood-data}
We used the following OOD datasets for ImageNet-1K:
\begin{itemize}
    \item iNaturalist \cite{van2018inaturalist}: iNaturalist is a dataset containing over 5,000 species of plants and animals. In \cite{huang2021mos}, a subset of 110 plant classes not present in ImageNet-1K was manually selected, and 10,000 images were randomly sampled from these classes to serve as OOD for ImageNet-1K.

    \item OpenImage-O \cite{wang2022vim}: This dataset consists of 17,632 samples from the OpenImage-v3 test set, which were manually verified as OOD with respect to ImageNet-1K.
    
    \item NINCO \cite{bitterwolf2023ninco}: NINCO consists of 5,879 samples from 64 classes, collected from various sources, including existing datasets and the internet. Each sample was manually verified to be truly OOD with respect to ImageNet-1K.

    \item Clean \cite{bitterwolf2023ninco}: Clean is a collection of 2,715 OOD images obtained from an analysis of 400 random samples drawn from 12 common OOD datasets (such as PLACES, Textures, Species, SSB-HARD, etc.), which were manually evaluated to determine whether they were truly OOD.
\end{itemize}

\subsection{Baselines Implementation Details}
\label{sec:supp-baseline-implementation}
For all baselines, we used the original source code from their GitHub repositories. The only exception is ZOC, where, following \cite{ming2022delving,jiang2024neglabel}, we upgraded the caption generator to BLIP (a more advanced model) to improve its effectiveness on ImageNet-1K; other implementation details for this method remain unchanged from the original paper.

\subsection{Randomness and Standard Deviations}
\label{sec:supp-randomness}
CSP involves randomness in generating modified conjugated class labels, where adjectives from WordNet are concatenated with a randomly selected term from a predefined list of 14 superclass labels. Our method builds on CSP and inherits this same source of randomness. To account for this, we repeated the experiments for both our method and CSP 10 times (using seeds 0 to 9). Additionally, CLIPN requires training an extra model on top of CLIP, and the authors have provided model snapshots from three runs in their GitHub repository. Other baselines do not involve any randomness. \cref{tab:main-std} reports the average and standard deviation for our method and CSP (based on the ten repetitions), as well as for CLIPN (based on the three model snapshots).

\subsection{Extended Analysis and Ablations}
\label{sec:supp-ablations}
\noindent \textbf{Ablation on the Size of the Initial Negative Label Set.}
For the initial set of negative labels, we relied on CSP, which, similar to NegLabel, selects the top $p = 15\%$ of words least related to the in-distribution labels as negative labels. \cref{tab:p} presents an ablation study evaluating different values of $p$ and comparing the performance of our method with CSP and NegLabel. Our method consistently outperforms both methods in average AUROC and FPR95 across all values of $p$. The lowest average FPR95 for our method is achieved at $p=40\%$, while CSP and NegLabel reach their optimal FPR95 at $p=60\%$ and $p=30\%$, respectively.

\noindent \textbf{Ablation on Different LLMs.} 
Our negative label filtering mechanism (NegFilter) leverages an LLM to identify proper nouns and subcategories. \cref{tab:llm} reports an ablation study on four mid-sized open-source LLMs. The results show that all LLMs yield nearly identical performance, suggesting that our method is not sensitive to the LLM choice.

\noindent \textbf{Ablation on Different CLIP Architectures.} 
In the main paper, following the literature, we adopted the CLIP ViT-B/16 model for our method and all baselines. In \cref{tab:clip}, we compare the performance of our method with CSP and NegLabel across different CLIP architectures. The results show that our method consistently outperforms both methods across all architectures by a significant margin.

\noindent \textbf{Ablation on the Design and Parameters of $S_{MM}$.}
In \cref{tab:smm-eq}, we present an ablation study on alternative designs for the $S_{MM}$ score function and different values of the $\alpha$ parameter. We also report an ablation on varying $k$ in \cref{tab:smm-k}. Notably, performance changes significantly from $k = 1$ to $k = 2$, while other $k$ values yield almost similar results.

\subsection{Evaluation Using Additional In-Distribution Datasets}
\label{sec:supp-other-in-distribution}
Following NegLabel and CSP, we also evaluate our method using other in-distribution datasets, including 
Stanford-Cars \cite{krause20133d}, CUB-200 \cite{wah2011caltech}, Oxford-Pet \cite{parkhi2012cats}, and Food-101 \cite{bossard2014food}, considering iNaturalist \cite{van2018inaturalist}, SUN \cite{xiao2010sun}, Places \cite{zhou2017places}, and Textures \cite{cimpoi2014describing} as OOD. \cref{tab:other-indist-data} compares our method with CSP and NegLabel. All methods achieve near-optimal performance across most datasets, with CSP and NegLabel slightly outperforming ours in some cases. Notably, these in-distribution datasets belong to narrow and distinct domains (cars, birds, pets, and food), making them easily distinguishable from the selected OOD datasets. As a result, all three methods exhibit similar performance.

\begin{table*}[h!]
  \centering
  \resizebox{\textwidth}{!}{
  \begin{tabular}{@{}lcccccccc@{}}
    \toprule
    \multirow{2}{*}{Methods} & \multicolumn{2}{c}{iNaturalist} & \multicolumn{2}{c}{OpenImage-O} & \multicolumn{2}{c}{Clean} & \multicolumn{2}{c}{NINCO} \\ \cmidrule(lr){2-3} \cmidrule(lr){4-5} \cmidrule(lr){6-7} \cmidrule(lr){8-9}
    & AUROC $\uparrow$ & FPR95 $\downarrow$ & AUROC $\uparrow$ & FPR95 $\downarrow$ & AUROC $\uparrow$ & FPR95 $\downarrow$ & AUROC $\uparrow$ & FPR95 $\downarrow$ \\
    \midrule
    CLIPN & 96.20 $\pm$ 0.41 & 19.11 $\pm$ 2.07 & 92.22 $\pm$ 0.42 & 30.36 $\pm$ 1.59 & 87.31 $\pm$ 0.46 & 41.56 $\pm$ 1.71 & 78.72 $\pm$ 1.19 & 66.51 $\pm$ 1.82 \\
    CSP  & \textbf{99.60} $\pm$ 0.00 & 1.54 $\pm$ 0.02 & 94.09 $\pm$ 0.03 & 28.94 $\pm$ 0.12 & 88.32 $\pm$ 0.05 & 38.75 $\pm$ 0.17 & 77.88 $\pm$ 0.17 & 68.65 $\pm$ 0.18 \\
    NegRefine (ours) & 99.57 $\pm$ 0.00 & \textbf{1.47} $\pm$ 0.03 & \textbf{95.00} $\pm$ 0.06 & \textbf{23.85} $\pm$ 0.16 & \textbf{90.65} $\pm$ 0.12 & \textbf{33.18} $\pm$ 0.37 & \textbf{81.92} $\pm$ 0.26 & \textbf{61.96} $\pm$ 0.41 \\
    \bottomrule
  \end{tabular}}
  \caption{Results including standard deviations for methods that involve randomness, corresponding to Table 1 in the main paper. For our method (NegRefine) and CSP, results are averaged over 10 different seeds. CLIPN results are based on the three model snapshots available in the authors' GitHub repository. Other baselines did not involve randomness. See \cref{sec:supp-randomness} for more details.}
  \label{tab:main-std}
\end{table*}

\begin{table*}
  \centering
  \resizebox{\textwidth}{!}{
  \begin{tabular}{@{}lccccccccc|cc@{}}
    \toprule
    \multirow{2}{*}{Methods} & \multirow{2}{*}{$p\%$} & \multicolumn{2}{c}{iNaturalist} & \multicolumn{2}{c}{OpenImage-O} & \multicolumn{2}{c}{Clean} & \multicolumn{2}{c}{NINCO} & \multicolumn{2}{|c}{Average} \\ \cmidrule(lr){3-4} \cmidrule(lr){5-6} \cmidrule(lr){7-8} \cmidrule(lr){9-10} \cmidrule(lr){11-12}
    && AUROC $\uparrow$ & FPR95 $\downarrow$ & AUROC $\uparrow$ & FPR95 $\downarrow$ & AUROC $\uparrow$ & FPR95 $\downarrow$ & AUROC $\uparrow$ & FPR95 $\downarrow$ & AUROC $\uparrow$ & FPR95 $\downarrow$ \\
    \midrule
    \multirow{10}{*}{NegLabel}
    & 15 & 99.52 & 1.85 & 93.74 & 28.62 & 86.79 & 41.22 & 77.30 & 68.70 & 89.34 & 35.10 \\
    & 20 & 99.41 & 2.30 & 93.95 & 27.83 & 86.72 & 40.85 & 76.78 & 68.32 & 89.22 & 34.82 \\
    & \cc 30 &\cc 99.20 &\cc 3.16 &\cc 94.17 &\cc 27.25 &\cc 86.67 &\cc 40.55 &\cc 76.61 &\cc 67.78 &\cc 89.16 &\cc 34.68 \\
    & 40 & 99.01 & 3.90 & 94.23 & 27.21 & 86.46 & 41.18 & 75.96 & 67.78 & 88.91 & 35.02 \\
    & 50 & 98.84 & 4.71 & 94.23 & 27.11 & 86.28 & 41.51 & 75.54 & 67.81 & 88.72 & 35.29 \\
    & 60 & 98.67 & 5.53 & 94.20 & 27.28 & 86.04 & 41.88 & 75.19 & 67.74 & 88.53 & 35.61 \\
    & 70 & 98.50 & 6.36 & 94.15 & 27.43 & 85.80 & 42.69 & 74.75 & 67.95 & 88.30 & 36.11 \\
    & 80 & 98.34 & 7.17 & 94.08 & 27.50 & 85.58 & 43.09 & 74.45 & 67.98 & 88.11 & 36.44 \\
    & 90 & 98.19 & 7.84 & 94.01 & 27.86 & 85.37 & 43.79 & 74.10 & 68.65 & 87.92 & 37.04 \\
    & 100 & 98.05 & 8.53 & 93.95 & 28.05 & 85.17 & 44.42 & 73.86 & 68.59 & 87.76 & 37.40 \\ \midrule
    
    \multirow{10}{*}{CSP}
    & 15 & 99.60 & 1.54 & 94.09 & 28.94 & 88.32 & 38.75 & 77.88 & 68.65 & 89.97 & 34.47 \\
    & 20 & 99.54 & 1.72 & 94.36 & 28.11 & 88.31 & 38.42 & 77.41 & 68.53 & 89.90 & 34.19 \\
    & 30 & 99.39 & 2.36 & 94.66 & 26.68 & 88.22 & 38.08 & 77.64 & 67.73 & 89.98 & 33.71 \\
    & 40 & 99.27 & 2.88 & 94.83 & 25.64 & 88.13 & 37.75 & 77.51 & 66.98 & 89.93 & 33.31 \\
    & 50 & 99.14 & 3.35 & 94.90 & 25.12 & 88.02 & 37.64 & 77.57 & 66.54 & 89.91 & 33.16 \\
    &\cc 60 &\cc 99.02 &\cc 3.96 &\cc 94.97 &\cc 24.48 &\cc 87.91 &\cc 37.20 &\cc 77.54 &\cc 66.26 &\cc 89.86 &\cc 32.98 \\
    & 70 & 98.90 & 4.47 & 95.01 & 24.18 & 87.78 & 37.53 & 77.47 & 66.23 & 89.79 & 33.10 \\
    & 80 & 98.79 & 5.05 & 95.02 & 24.02 & 87.63 & 37.64 & 77.35 & 65.75 & 89.70 & 33.12 \\
    & 90 & 98.70 & 5.50 & 95.01 & 24.15 & 87.46 & 38.42 & 77.13 & 65.58 & 89.57 & 33.41 \\
    & 100 & 98.60 & 5.85 & 94.98 & 24.11 & 87.20 & 38.71 & 76.75 & 65.28 & 89.38 & 33.49 \\ \midrule
    
    \multirow{10}{*}{NegRefine}
    & 15 & 99.57 & 1.47 & 95.00 & 23.85 & 90.65 & 33.18 & 81.92 & 61.96 & 91.79 & 30.12 \\
    & 20 & 99.51 & 1.80 & 95.08 & 23.44 & 90.45 & 33.26 & 81.76 & 61.81 & 91.70 & 30.08 \\
    & 30 & 99.37 & 2.28 & 95.26 & 22.97 & 90.56 & 32.97 & 81.95 & 61.11 & 91.78 & 29.83 \\
    &\cc 40 &\cc 99.21 &\cc 2.77 &\cc 95.28 &\cc 22.86 &\cc 90.29 &\cc 32.82 &\cc 81.84 &\cc 60.41 &\cc 91.66 &\cc 29.71\\
    & 50 & 99.07 & 3.47 & 95.31 & 22.74 & 90.01 & 33.41 & 81.35 & 60.21 & 91.44 & 29.96 \\
    & 60 & 98.91 & 4.05 & 95.30 & 22.56 & 89.74 & 33.70 & 81.19 & 60.39 & 91.28 & 30.18 \\
    & 70 & 98.78 & 4.49 & 95.27 & 22.64 & 89.42 & 34.00 & 80.57 & 60.80 & 91.01 & 30.48 \\
    & 80 & 98.64 & 4.95 & 95.20 & 23.01 & 89.11 & 34.40 & 80.14 & 60.96 & 90.77 & 30.83 \\
    & 90 & 98.47 & 5.52 & 95.05 & 23.38 & 88.71 & 35.06 & 79.51 & 60.99 & 90.43 & 31.24\\
    & 100 & 98.27 & 6.20 & 94.78 & 24.38 & 88.15 & 36.28 & 78.55 & 61.77 & 89.94 & 32.16 \\
    \bottomrule
  \end{tabular}}
  \caption{Ablation study on different values of the initial negative label selection percentage $p$ using the ImageNet-1K in-distribution benchmark. For each method, the value of $p$ that achieves the best average FPR95 is highlighted in gray. Across all values of $p$, our method (NegRefine) consistently outperforms CSP and NegLabel in both average AUROC and FPR95.}
  \label{tab:p}
\end{table*}

\begin{table*}
  \centering
  \resizebox{\textwidth}{!}{
  \begin{tabular}{@{}lcccccccc|cc@{}}
    \toprule
    \multirow{2}{*}{LLM} & \multicolumn{2}{c}{iNaturalist} & \multicolumn{2}{c}{OpenImage-O} & \multicolumn{2}{c}{Clean} & \multicolumn{2}{c}{NINCO} & \multicolumn{2}{|c}{Average} \\ \cmidrule(lr){2-3} \cmidrule(lr){4-5} \cmidrule(lr){6-7} \cmidrule(lr){8-9} \cmidrule(lr){10-11}
    & AUROC $\uparrow$ & FPR95 $\downarrow$ & AUROC $\uparrow$ & FPR95 $\downarrow$ & AUROC $\uparrow$ & FPR95 $\downarrow$ & AUROC $\uparrow$ & FPR95 $\downarrow$ & AUROC $\uparrow$ & FPR95 $\downarrow$ \\
    \midrule
    Qwen2.5-14B-Instruct & 99.57 & 1.47 & \textbf{95.00} & \textbf{23.85} & \textbf{90.65} & \textbf{33.18} & 81.92 & 61.96 & 91.79 & 30.12 \\
    Qwen2.5-7B-Instruct-1M & 99.55 & 1.61 & 94.98 & 23.94 & 90.60 & 33.63 & 81.83 & 61.86 & 91.74 & 30.26 \\
    Mistral-7B-Instruct-v0.2 & 99.60 & 1.36 & 94.82 & 24.34 & 90.63 & 33.37 & \textbf{82.92} & \textbf{61.14} & \textbf{91.99} & 30.05 \\
    Meta-Llama-3-8B-Instruct & \textbf{99.61} & \textbf{1.32} & 94.92 & 24.04 & 90.56 & 33.48 & 82.61 & 61.31 & 91.93 & \textbf{30.04}  \\
    \bottomrule
  \end{tabular}}
  \caption{Ablation study on different LLMs for negative label filtering using the ImageNet-1K in-distribution benchmark.}
  \label{tab:llm}
\end{table*}

\begin{table*}
  \centering
  \resizebox{\textwidth}{!}{
  \begin{tabular}{@{}lccccccccc|cc@{}}
    \toprule
    \multirow{2}{*}{Architecture} & \multirow{2}{*}{Method} & \multicolumn{2}{c}{iNaturalist} & \multicolumn{2}{c}{OpenImage-O} & \multicolumn{2}{c}{Clean} & \multicolumn{2}{c}{NINCO} & \multicolumn{2}{|c}{Average} \\ \cmidrule(lr){3-4} \cmidrule(lr){5-6} \cmidrule(lr){7-8} \cmidrule(lr){9-10} \cmidrule(lr){11-12}
    & & AUROC $\uparrow$ & FPR95 $\downarrow$ & AUROC $\uparrow$ & FPR95 $\downarrow$ & AUROC $\uparrow$ & FPR95 $\downarrow$ & AUROC $\uparrow$ & FPR95 $\downarrow$ & AUROC $\uparrow$ & FPR95 $\downarrow$ \\
    \midrule
    \multirow{2}{*}{ResNet50}
    & NegLabel & 99.24 & 2.88 & 92.00 & 33.35 & 84.45 & 45.52 & 74.20 & 73.53 & 87.47 & 38.82 \\
    & CSP & \textbf{99.46} & \textbf{1.95} & 91.83 & 33.91 & 86.08 & 42.06 & 75.53 & 72.86 & 88.22 & 37.70 \\
    & NegRefine & 99.38 & 2.30 & \textbf{92.57} & \textbf{31.05} & \textbf{88.26} & \textbf{38.97} & \textbf{80.73} & \textbf{65.62} & \textbf{90.23} & \textbf{34.48} \\ \midrule
    \multirow{2}{*}{ResNet101}
    & NegLabel & 99.27 & 3.11 & 90.92 & 37.41 & 84.83 & 46.34 & 76.80 & 72.93 & 87.95 & 39.95 \\
    & CSP & 99.47 & 2.04 & 91.90 & 35.58 & 85.81 & 43.43 & 75.92 & 72.25 & 88.28 & 38.33  \\
    & NegRefine & \textbf{99.59} & \textbf{1.49} & \textbf{93.75} & \textbf{27.52} & \textbf{88.43} & \textbf{38.23} & \textbf{80.17} & \textbf{66.06} & \textbf{90.48} & \textbf{33.33} \\ \midrule
    \multirow{2}{*}{ResNet50x4}
    & NegLabel & 99.45 & 2.27 & 91.93 & 34.83 & 86.28 & 43.24 & 78.33 & 69.96 & 89.00 & 37.58 \\
    & CSP & 99.65 & 1.48 & 93.38 & 31.62 & 87.84 & 39.01 & 79.14 & 68.25 & 90.00 & 35.09 \\
    & NegRefine & \textbf{99.66} & \textbf{1.25} & \textbf{94.32} & \textbf{26.16} & \textbf{89.75} & \textbf{35.51} & \textbf{82.27} & \textbf{61.53} & \textbf{91.50} & \textbf{31.11} \\ \midrule
    \multirow{2}{*}{ResNet50x16}
    & NegLabel & 99.48 & 2.00 & 92.50 & 33.82 & 88.12 & 40.41 & 80.56 & 65.57 & 90.17 & 35.45 \\
    & CSP & \textbf{99.68} & \textbf{1.25} & 93.91 & 30.47 & 89.92 & 38.20 & 82.20 & 63.22 & 91.43 & 33.28 \\
    & NegRefine & 99.60 & 1.42 & \textbf{94.68} & \textbf{25.59} & \textbf{91.19} & \textbf{34.40} & \textbf{84.18} & \textbf{57.18} & \textbf{92.41} & \textbf{29.65} \\ \midrule
    \multirow{2}{*}{ResNet50x64}
    & NegLabel & 99.63 & 1.46 & 93.68 & 30.58 & 90.15 & 37.42 & 86.98 & 54.41 & 92.61 & 30.97 \\
    & CSP & \textbf{99.69} & \textbf{1.19} & 93.95 & 30.21 & 91.48 & 33.92 & 87.18 & 52.94 & 93.08 & 29.57 \\
    & NegRefine & 99.60 & 1.41 & \textbf{94.52} & \textbf{26.45} & \textbf{92.14} & \textbf{30.90} & \textbf{88.26} & \textbf{47.33} & \textbf{93.63} & \textbf{26.52} \\ \midrule
    \multirow{2}{*}{ViT-B/32}
    & NegLabel & 99.11 & 3.73 & 92.87 & 31.26 & 85.20 & 44.16 & 75.30 & 71.45 & 88.12 & 37.65 \\
    & CSP & \textbf{99.46} & 2.37 & 93.37 & 31.01 & 87.42 & 40.15 & 76.96 & 70.62 & 89.30 & 36.04 \\
    & NegRefine & 99.45 & \textbf{2.20} & \textbf{94.22} & \textbf{26.11} & \textbf{89.90} & \textbf{35.47} & \textbf{82.38} & \textbf{60.70} & \textbf{91.49} & \textbf{31.12} \\ \midrule
    \multirow{2}{*}{ViT-B/16}
    & NegLabel & 99.49 & 1.91 & 93.74 & 28.62 & 86.79 & 41.22 & 77.30 & 68.70 & 89.33 & 35.11 \\
    & CSP & \textbf{99.60} & 1.54 & 94.09 & 28.94 & 88.32 & 38.75 & 77.88 & 68.65 & 89.97 & 34.47 \\
    & NegRefine & 99.57 & \textbf{1.47} & \textbf{95.00} & \textbf{23.85} & \textbf{90.65} & \textbf{33.18} & \textbf{81.92} & \textbf{61.96} & \textbf{91.79} & \textbf{30.12} \\ \midrule
    \multirow{2}{*}{ViT-L/14}
    & NegLabel & 99.53 & 1.77 & 94.26 & 27.55 & 89.09 & 38.38 & 81.98 & 62.59 & 91.22 & 32.57 \\
    & CSP & \textbf{99.72} & \textbf{1.21} & 95.02 & 25.59 & 91.52 & 34.22 & 84.94 & 57.20 & 92.80 & 29.55 \\
    & NegRefine & 99.67 & 1.25 & \textbf{95.96} & \textbf{19.92} & \textbf{92.74} & \textbf{28.43} & \textbf{87.62} & \textbf{47.40} & \textbf{94.00} & \textbf{24.25} \\ \midrule
    \multirow{2}{*}{ViT-L/14-336px}
    & NegLabel & 99.67 & 1.31 & 94.26 & 27.28 & 89.64 & 36.87 & 83.70 & 61.06 & 91.82 & 31.63 \\
    & CSP & \textbf{99.79} & \textbf{0.86} & 94.90 & 26.06 & 91.99 & 32.19 & 86.38 & 55.90 & 93.27 & 28.75 \\
    & NegRefine & 99.71 & 1.01 & \textbf{95.73} & \textbf{20.46} & \textbf{93.16} & \textbf{26.92} & \textbf{88.81} & \textbf{44.68} & \textbf{94.35} & \textbf{23.27} \\
    \bottomrule
  \end{tabular}}
  \caption{Ablation study comparing our method (NegRefine) with CSP and NegLabel across different CLIP architectures using the ImageNet-1K in-distribution benchmark.}
  \label{tab:clip}
\end{table*}

\begin{table*}
  \centering
  \resizebox{\textwidth}{!}{
  \begin{tabular}{@{}cccccccccc|cc@{}}
    \toprule
    && \multicolumn{8}{c}{OOD Datasets} & \\
    \multirow{2}{*}{$S_{MM}$ Score} & \multirow{2}{*}{$\alpha$} & \multicolumn{2}{c}{iNaturalist} & \multicolumn{2}{c}{OpenImage-O} & \multicolumn{2}{c}{Clean} & \multicolumn{2}{c}{NINCO} & \multicolumn{2}{|c}{Average} \\ \cmidrule(lr){3-4} \cmidrule(lr){5-6} \cmidrule(lr){7-8} \cmidrule(lr){9-10} \cmidrule(lr){11-12}
    & & AUROC $\uparrow$ & FPR95 $\downarrow$ & AUROC $\uparrow$ & FPR95 $\downarrow$ & AUROC $\uparrow$ & FPR95 $\downarrow$ & AUROC $\uparrow$ & FPR95 $\downarrow$ & AUROC $\uparrow$ & FPR95 $\downarrow$ \\
    \midrule

    \multirow{4}{*}{$ \max_{i, j} \; \left( \operatorname{sim}( x, t_{i,j})/\tau - \operatorname{sim}(x, \tilde{y}_j)/\tau \right) $}
    & 1 & 98.10 & 8.51 & 92.57 & 31.85 & 89.60 & 37.79 & 81.63 & 62.93 & 90.47 & 35.27 \\
    & 2 & 97.71 & 10.35 & 92.30 & 33.10 & 89.39 & 38.60 & 81.55 & 63.34 & 90.24 & 36.35 \\
    & 3 & 97.54 & 11.10 & 92.20 & 33.64 & 89.31 & 39.01 & 81.53 & 63.42 & 90.15 & 36.79 \\
    & 4 & 97.45 & 11.55 & 92.14 & 33.96 & 89.27 & 39.30 & 81.51 & 63.39 & 90.09 & 37.05 \\ \midrule

    \multirow{4}{*}{$ \max_{i, j} \; \frac{ \operatorname{sim}( x, t_{i,j})/\tau }{ \operatorname{sim}( x, t_{i,j})/\tau + \operatorname{sim}(x, \tilde{y}_j)/\tau } $}
    & 1 & \textbf{99.65} & 1.26 & 94.85 & 24.16 & 90.50 & 34.66 & 80.76 & 65.65 & 91.44 & 31.43 \\
    & 2 & \textbf{99.65} & \textbf{1.24} & 94.66 & 23.65 & 90.68 & 33.92 & 81.11 & 64.78 & 91.53 & 30.90 \\
    & 3 & 99.63 & 1.29 & 94.49 & \textbf{23.62} & \textbf{90.71} & 33.70 & 81.26 & 64.15 & 91.52 & 30.69 \\
    & 4 & 99.62 & 1.32 & 94.34 & 23.63 & 90.69 & 33.15 & 81.33 & 63.54 & 91.49 & 30.41 \\ \midrule
    
    \multirow{4}{*}{$ \max_{i, j} \; \frac{ e^{\operatorname{sim}( x, t_{i,j})/\tau} }{ e^{\operatorname{sim}( x, t_{i,j})/\tau} + e^{\operatorname{sim}(x, \tilde{y}_j)/\tau} } $}
    & 1 &  99.62 & 1.38 & \textbf{95.07} & 23.63 & 90.39 & 33.63 & 80.77 & 63.78 & 91.46 & 30.61 \\
    &\cc 2 &\cc  99.57 &\cc 1.47 &\cc 95.00 &\cc 23.85 &\cc 90.65 &\cc 33.18 &\cc \textbf{81.92} &\cc \textbf{61.96} &\cc \textbf{91.79} &\cc \textbf{30.12} \\ 
    & 3 &  99.50 & 1.67 & 94.85 & 24.30 & 90.65 & \textbf{33.04} & 81.49 & 62.61 & 91.62 & 30.41 \\
    & 4 &  99.43 & 1.76 & 94.72 & 24.85 & 90.66 & 33.19 & 81.63 & 62.11 & 91.61 & 30.48 \\ \midrule

    \multirow{4}{*}{$ \text{avg}_{i, j} \; \frac{ e^{\operatorname{sim}( x, t_{i,j})/\tau} }{ e^{\operatorname{sim}( x, t_{i,j})/\tau} + e^{\operatorname{sim}(x, \tilde{y}_j)/\tau} } $}
    & 1 &  98.95 & 3.79 & 91.01 & 32.12 & 87.81 & 39.04 & 72.67 & 70.14 & 87.61 & 36.27 \\
    & 2 &  98.36 & 6.21 & 89.85 & 36.06 & 86.96 & 41.44 & 71.93 & 71.86 & 86.77 & 38.89 \\
    & 3 &  97.86 & 8.48 & 89.23 & 38.22 & 86.48 & 43.28 & 71.58 & 73.07 & 86.29 & 40.76 \\
    & 4 &  97.46 & 10.63 & 88.83 & 39.75 & 86.16 & 44.83 & 71.38 & 73.55 & 85.96 & 42.19 \\ \midrule

    \multirow{4}{*}{$ \sum_{i, j} \; \frac{ e^{\operatorname{sim}( x, t_{i,j})/\tau} }{ e^{\operatorname{sim}( x, t_{i,j})/\tau} + e^{\operatorname{sim}(x, \tilde{y}_j)/\tau} } $}
    & 1 & 95.41 & 20.08 & 87.31 & 45.84 & 84.92 & 49.65 & 70.69 & 76.32 & 84.58 & 47.97 \\
    & 2 & 95.08 & 21.73 & 87.11 & 46.86 & 84.75 & 50.68 & 70.60 & 76.73 & 84.38 & 49.00 \\
    & 3 & 94.97 & 22.24 & 87.04 & 47.16 & 84.69 & 51.16 & 70.57 & 76.83 & 84.32 & 49.35 \\
    & 4 & 94.91 & 22.48 & 87.00 & 47.31 & 84.66 & 51.31 & 70.56 & 76.86 & 84.28 & 49.49 \\

    \bottomrule
  \end{tabular}}
  \caption{Ablation study on different design choices for the $S_{MM}$ score function across various $\alpha$ values. The configuration adopted in the main paper is highlighted in gray.}
  \label{tab:smm-eq}
\end{table*}

\begin{table*}
  \centering
  \resizebox{\textwidth}{!}{
  \begin{tabular}{@{}ccccccccc|cc@{}}
    \toprule
    && \multicolumn{8}{c}{OOD Datasets} & \\
    \multirow{2}{*}{$k$} & \multicolumn{2}{c}{iNaturalist} & \multicolumn{2}{c}{OpenImage-O} & \multicolumn{2}{c}{Clean} & \multicolumn{2}{c}{NINCO} & \multicolumn{2}{|c}{Average} \\ \cmidrule(lr){2-3} \cmidrule(lr){4-5} \cmidrule(lr){6-7} \cmidrule(lr){8-9} \cmidrule(lr){10-11}
    & AUROC $\uparrow$ & FPR95 $\downarrow$ & AUROC $\uparrow$ & FPR95 $\downarrow$ & AUROC $\uparrow$ & FPR95 $\downarrow$ & AUROC $\uparrow$ & FPR95 $\downarrow$ & AUROC $\uparrow$ & FPR95 $\downarrow$ \\
    \midrule
    1 & 98.40 & 7.63 & 92.03 & 34.87 & 88.84 & 40.66 & 80.71 & 64.29 & 89.99 & 36.86 \\
    2 & 99.21 & 2.82 & 93.94 & 28.19 & 90.24 & 35.43 & 82.11 & \textbf{61.31} & 91.38 & 31.94 \\
    3 & 99.43 & 2.00 & 94.59 & 25.58 & 90.53 & \textbf{32.93} & 82.10 & 61.87 & 91.66 & 30.59 \\
    4 & 99.53 & 1.71 & 94.94 & 24.30 & \textbf{90.71} & 32.97 & \textbf{82.24} & 61.67 & \textbf{91.86} & 30.16 \\
    \rowcolor{gray!40} 5 & 99.57 & 1.47 & 95.00 & 23.85 & 90.65 & 33.18 & 81.92 & 61.96 & 91.79 & \textbf{30.12} \\
    6 & 99.59 & 1.50 & 95.12 & 23.73 & 90.61 & 33.30 & 82.11 & 62.45 & \textbf{91.86} & 30.25 \\
    7 & 99.61 & 1.46 & 95.17 & 23.54 & 90.59 & 33.19 & 82.01 & 62.59 & 91.84 & 30.20 \\
    8 & 99.61 & 1.40 & 95.20 & 23.22 & 90.51 & 33.19 & 81.89 & 62.71 & 91.80 & 30.13 \\
    9 & 99.63 & 1.37 & 95.22 & 23.22 & 90.44 & 33.48 & 81.81 & 62.93 & 91.77 & 30.25 \\
    10 & 99.63 & 1.38 & \textbf{95.23} & 23.33 & 90.42 & 33.30 & 81.75 & 63.24 & 91.76 & 30.31 \\
    15 & \textbf{99.65} & 1.35 & \textbf{95.23} & \textbf{23.17} & 90.22 & 33.70 & 81.43 & 64.10 & 91.63 & 30.58 \\
    20 & \textbf{99.65} & \textbf{1.32} & 95.21 & 23.34 & 90.03 & 33.78 & 81.20 & 64.09 & 91.52 & 30.63 \\
    \bottomrule
  \end{tabular}}
  \caption{Ablation study on different values of $k$ (the number of top in-distribution and negative labels used to create multi-matching texts). In the paper, we adopted $k = 5$ (highlighted in gray).}
  \label{tab:smm-k}
\end{table*}

\begin{table*}
  \centering
  \resizebox{\textwidth}{!}{
  \begin{tabular}{@{}lccccccccc|cc@{}}
    \toprule
    && \multicolumn{8}{c}{OOD Datasets} & \\
    \multirow{2}{*}{In-distribution data} & \multirow{2}{*}{Method} & \multicolumn{2}{c}{iNaturalist} & \multicolumn{2}{c}{SUN} & \multicolumn{2}{c}{Places} & \multicolumn{2}{c}{Textures} & \multicolumn{2}{|c}{Average} \\ \cmidrule(lr){3-4} \cmidrule(lr){5-6} \cmidrule(lr){7-8} \cmidrule(lr){9-10} \cmidrule(lr){11-12}
    & & AUROC $\uparrow$ & FPR95 $\downarrow$ & AUROC $\uparrow$ & FPR95 $\downarrow$ & AUROC $\uparrow$ & FPR95 $\downarrow$ & AUROC $\uparrow$ & FPR95 $\downarrow$ & AUROC $\uparrow$ & FPR95 $\downarrow$ \\
    \midrule
    \multirow{3}{*}{Stanford-Cars}
    & NegLabel & 99.99 & 0.01 & 99.99 & 0.01 & \textbf{99.99} & 0.03 & 99.99 & 0.01 & 99.99 & 0.02 \\
    & CSP & \textbf{100.0} & \textbf{0.00} & \textbf{100.0} & \textbf{0.00} & \textbf{99.99} & \textbf{0.02} & \textbf{100.0} & \textbf{0.00} & \textbf{100.0} & \textbf{0.01} \\
    & NegRefine & \textbf{100.0} & \textbf{0.00} & \textbf{100.0} & 0.01 & \textbf{99.99} & 0.07 & \textbf{100.0} & \textbf{0.00} & \textbf{100.0} & 0.02 \\ \midrule
    \multirow{3}{*}{CUB-200}
    & NegLabel & \textbf{99.96} & 0.18 & \textbf{99.99} & \textbf{0.02} & \textbf{99.90} & \textbf{0.33} & 99.99 & 0.01 & \textbf{99.96} & \textbf{0.14} \\
    & CSP & \textbf{99.96} & \textbf{0.16} & \textbf{99.99} & 0.03 & 99.88 & 0.37 & \textbf{100.0} & \textbf{0.00} & \textbf{99.96} & \textbf{0.14} \\
    & NegRefine & 99.92 & 0.19 & 99.95 & 0.14 & 99.68 & 1.04 & 99.99 & 0.02 & 99.89 & 0.35 \\ \midrule
    \multirow{3}{*}{Oxford-Pet}
    & NegLabel & 99.99 & 0.01 & 99.99 & 0.02 & \textbf{99.96} & \textbf{0.17} & \textbf{99.97} & \textbf{0.11} & \textbf{99.98} & \textbf{0.08} \\
    & CSP & \textbf{100.0} & \textbf{0.00} & \textbf{100.0} & \textbf{0.00} & \textbf{99.96} & 0.21 & \textbf{99.97} & 0.14 & \textbf{99.98} & 0.09 \\
    & NegRefine & \textbf{100.0} & \textbf{0.00} & 99.99 & 0.01 & 99.95 & 0.24 & 99.96 & 0.12 & 99.97 & 0.09 \\ \midrule
    \multirow{3}{*}{Food-101}
    & NegLabel & 99.99 & 0.01 & 99.99 & 0.01 & \textbf{99.99} & 0.01 & 99.60 & 1.61 & 99.89 & 0.41 \\
    & CSP & \textbf{100.0} & \textbf{0.00} & \textbf{100.0} & \textbf{0.00} & \textbf{99.99} & 0.01 & 99.63 & 1.40 & \textbf{99.91} & 0.35 \\
    & NegRefine & 99.99 & \textbf{0.00} & 99.98 & \textbf{0.00} & 99.98 & \textbf{0.00} & \textbf{99.68} & \textbf{1.37} & \textbf{99.91} & \textbf{0.34} \\ 
    \bottomrule
  \end{tabular}}
  \caption{OOD detection performance using other in-distribution datasets.}
  \label{tab:other-indist-data}
\end{table*}

\end{document}